\pdfoutput=1
\documentclass[11pt]{article}

\usepackage[]{acl}

\usepackage{times}
\usepackage{latexsym}
\usepackage{tabularx}
\usepackage{multirow}
\usepackage{pgfplots}
\usepackage{tikz}
\usepackage{pgfplots}
\usepackage{pgf-pie}
\usepackage{subcaption}
\usepackage{fixltx2e}
\usepackage{hyperref}
\usepackage{tablefootnote}
\usepackage{svg}
\usepackage{adjustbox}
\usepackage{hyperref}     
\usepackage{url}        
\usepackage{booktabs}      
\usepackage{amsfonts}     
\usepackage{nicefrac}      
\usepackage{microtype}      
\usepackage{xcolor}         
\usepackage{tabularx}
\usepackage{graphicx}
\usepackage{amsmath}

\definecolor{color1}{RGB}{159,168,218}
\definecolor{color2}{RGB}{173,201,198}
\definecolor{color3}{RGB}{207,216,220}
\definecolor{color4}{RGB}{247,194,183}
\definecolor{color5}{RGB}{187,230,190}
\definecolor{color6}{RGB}{187,230,222}
\definecolor{darkgreen}{rgb}{0,0.5,0}
\definecolor{fancy_color1}{RGB}{51,75,127}
\definecolor{fancy_color2}{RGB}{180,75,144}
\definecolor{pie_pink}{HTML}{feaba7}
\definecolor{pie_rose}{HTML}{fbe1ed}
\definecolor{pie_purple}{HTML}{dbd1ed}
\definecolor{pie_green}{HTML}{adeaad}
\definecolor{pie_orange}{HTML}{ffcd88}
\definecolor{pie_gray}{HTML}{dde7eb}
\definecolor{orchid}{RGB}{218, 112, 214}

\usepackage[T1]{fontenc}
\usepackage[utf8]{inputenc}

\usepackage{microtype}

\title{Evaluating Large Language Models on Wikipedia-Style Survey Generation}

\author{Fan Gao$^{1, 9}$, Hang Jiang$^2$, Rui Yang$^3$, Qingcheng Zeng$^4$, Jinghui Lu$^6$, \\
\textbf{Moritz Blum$^5$, Dairui Liu$^7$, Tianwei She$^8$, Yuang Jiang$^6$, Irene Li$^{1,6}$} \\
$^1$University of Tokyo, $^2$MIT Center for Constructive Communication, \\$^3$National University of Singapore, 
$^4$Northwestern University, $^5$Bielefeld University, \\
$^6$Smartor.me,
$^7$University College Dublin, $^8$Moveworks, $^9$Tokyo Institute of Technology\\
\small{\texttt{fangao0802@g.ecc.u-tokyo.ac.jp,} 
\texttt{ireneli@ds.itc.u-tokyo.ac.jp}}}

\begin{document}
\maketitle
\begin{abstract}

Educational materials such as survey articles in specialized fields like computer science traditionally require tremendous expert inputs and are therefore expensive to create and update. Recently, Large Language Models (LLMs) have achieved significant success across various general tasks. However, their effectiveness and limitations in the education domain are yet to be fully explored. In this work, we examine the proficiency of LLMs in generating succinct survey articles specific to the niche field of NLP in computer science, focusing on a curated list of 99 topics. Automated benchmarks reveal that GPT-4 surpasses its predecessors, inluding GPT-3.5, PaLM2, and LLaMa2 by margins ranging from 2\% to 20\% in comparison to the established ground truth. We compare both human and GPT-based evaluation scores and provide in-depth analysis. While our findings suggest that GPT-created surveys are more contemporary and accessible than human-authored ones, certain limitations were observed. Notably, GPT-4, despite often delivering outstanding content, occasionally exhibited lapses like missing details or factual errors. At last, we compared the rating behavior between humans and GPT-4 and found systematic bias in using GPT evaluation.
% In this study, we assess the capability LLMs to generate concise survey articles within the specialized NLP field in computer science, focusing on 100 chosen topics. Automated evaluations indicate that GPT-4 outperforms GPT-3.5, PaLM2, and LLaMa2 when benchmarked against the ground truth. Additionally, we conduct comprehensive human evaluations with a total of seven human experts, offering thoughtful insights on GPT-generated surveys. We find that these surveys are more up-to-date and accessible than human-written gold truth. However, case studies highlight that while GPT-4 often produces impressive results, failures, such as incomplete information and factual inaccuracies, occur. 

\end{abstract}

\begin{figure*}[h!]
  \centering
  \adjustbox{width=\textwidth}{\includegraphics{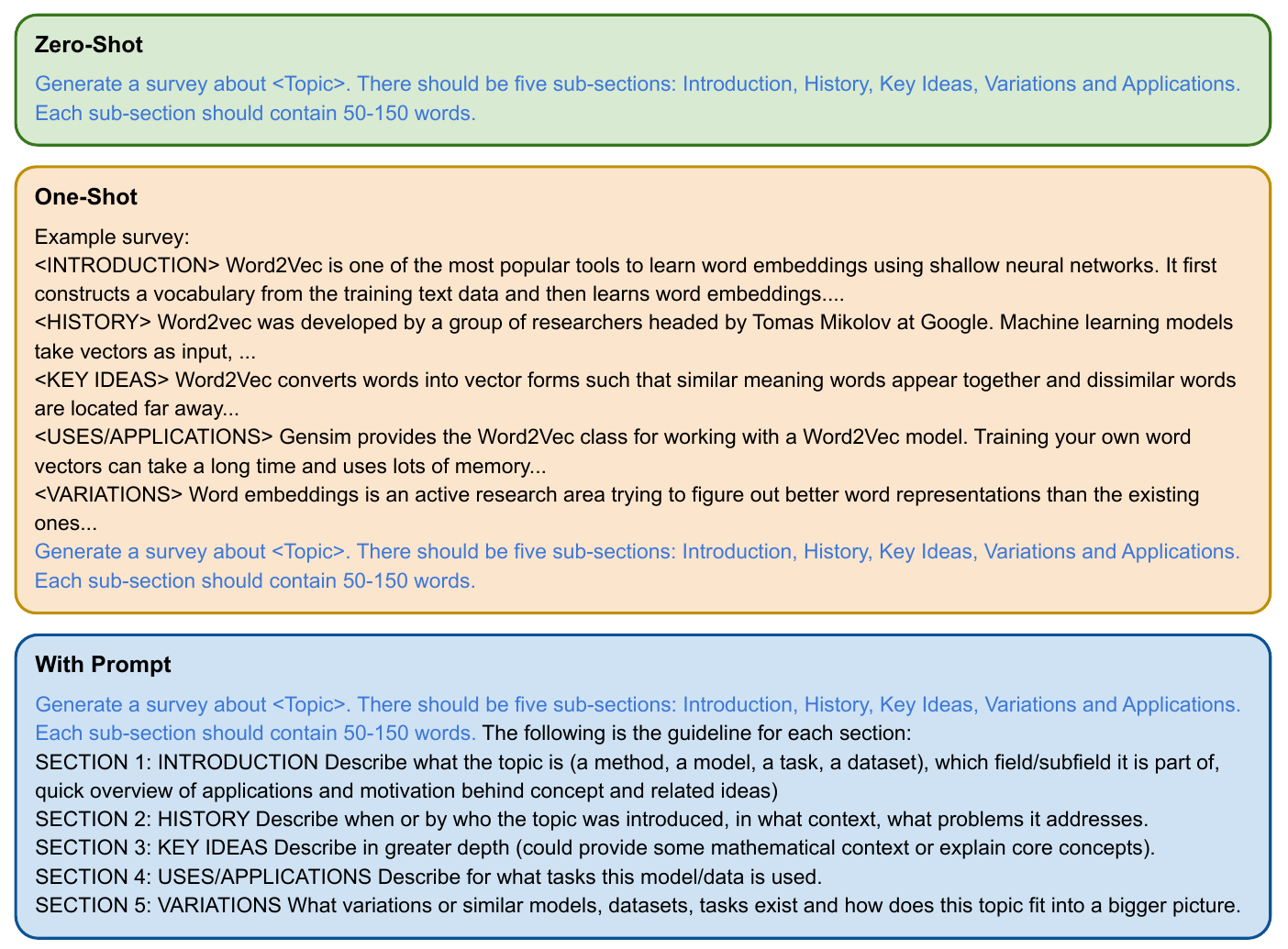}}
  \caption{The three main prompt types we compared. We eliminated some text in the one-shot setting, which is the ground truth from the survey of \texttt{Word2Vec}. }
  \label{tab:prompt}
\vspace{-2mm}
\end{figure*}

\section{Introduction}

Recently, large language models (LLMs) have attracted significant attention due to their strong performance on general natural language processing (NLP) tasks \cite{shaib2023summarizing,Feng2022DiffuserET}. Especially, the GPT family \cite{brown2020language} shows great ability in various applications. While it has been demonstrated that they perform well in many general tasks, their effectiveness in domain-specific tasks continues to be under scrutiny \cite{tian2023opportunities}. Specifically, the text produced by LLMs can sometimes exhibit issues like creating false information and hallucination \cite{zhao2023llm,Yang2024KGRankEL}.

In the context of scientific education, automatic survey generation aims to employ machine learning or NLP techniques to create a structured overview of a specific concept \cite{Sun2022AutomaticSG, li-etal-2022-surfer100,Yang2024LeveragingLL}. Automating this process not only alleviates the manual effort but also ensures timely updates at a reduced cost. A common approach involves an initial information retrieval phase to select pertinent documents or sentences based on the query topic. This is followed by a summarization or simplification phase to produce the final survey \cite{Jha2013ASF,li-etal-2022-surfer100}. While LLMs have the potential to be an alternative method for writing scientific surveys, their effectiveness and limitations are not yet thoroughly investigated. 

Existing work focuses on applying LLMs to similar scenarios, including aiding scientific writing \cite{shen2023beyond, altmae2023artificial}, question-answering with scientific papers \cite{Tahri2022OnTP}, writing paper reviews \cite{Liang2023CanLL}, and answering quiz or exam questions \cite{Song2023NLPBenchEL,Wang2023SciBenchEC}. This study pushes the boundary of this research area as the first to evaluate the capability of LLMs in generating education surveys within the scientific domain of NLP \cite{li-etal-2022-surfer100}. Our primary objective is to understand whether LLMs can be used to explain concepts in a more structured manner. To this end, we aim to answer the following research questions (RQs):
\vspace{-3mm}
\begin{itemize}
    \item \textbf{RQ1}: How proficient are LLMs in generating survey articles on NLP concepts?
    \vspace{-2mm}
    \item \textbf{RQ2}: Can LLMs emulate human judgment when provided with specific criteria? 
    \vspace{-2mm}
    \item \textbf{RQ3}: Do LLMs introduce a noticeable bias in evaluating machine-generated texts compared to human-written texts?
\end{itemize}
\vspace{-3mm}
We empirically conduct experiments on LLaMa2 \cite{touvron2023llama}, PaLM2 \cite{anil2023palm}, GPT-3.5 \cite{brown2020language} and GPT-4 \cite{openai2023gpt4} across four different settings. Furthermore, we engage human experts to provide a qualitative dimension, ensuring that our results not only reflect the technical performance but also incorporate subjective human perspectives. We release the LLMs-generated surveys of all these works\footnote{\url{https://github.com/astridesa/EDULLM/tree/master}}.

\section{Method}

\subsection{Dataset}
We adopt the Surfer100 dataset \cite{li-etal-2022-surfer100}, which contains 100 manually written Wikipedia-style articles on NLP concepts. Each survey is structured into five sections: \textit{Introduction}, \textit{History}, \textit{Key Ideas}, \textit{Uses/Applications}, and \textit{Variations}, with each section containing between 50 and 150 words. 

\subsection{Experiment Setup}
We compare three settings: \textbf{zero-shot (ZS)}, \textbf{one-shot (OS)} and \textbf{description prompt (DP)}. For zero-shot, we directly ask the model to generate the article by providing the following prompt: \textit{Generate a survey about <Topic>. There should be five sub-sections: Introduction, History, Key Ideas, Variations and Applications. Each subsection should contain 50-150 words.} For the one-shot setting, we add a ground-truth article of \texttt{Word2Vec} as the sample survey; for the description prompt setting, we add a detailed description to each section explaining what should be included. For example, \textit{SECTION 1: INTRODUCTION
Describe what the topic is (a method, a model, a task, a dataset), which field/subfield it is part of, quick overview of applications and motivation behind the concept and related ideas)}. To further enrich the provided information, we also introduce a combination of one-shot and description prompt \textbf{(OSP)}. The full prompt is shown in Figure~\ref{tab:prompt}. By employing a single ground truth for one-shot learning, we generate 99 surveys per setting. Moreover, we evaluate a special retrieval augmented generation (RAG) ~\cite{gao2023retrieval, ram2023context, yang2023leandojo, aksitov2023characterizing} setting (denoted as \textbf{OS+IR)} which enables GPT-4 link to Wikipedia pages and access to web data. We elaborate more experimental details in Appendix~\ref{app:wiki}

\begin{table*}[t]
\centering
\small
\renewcommand{\arraystretch}{0.8}
\begin{tabular}{lccccccccc}
\toprule
\multirow{2}{*}{Method} & \multicolumn{3}{c}{ROUGE} & \multicolumn{3}{c}{BERTScore} & \multirow{2}{*}{MoverScore} & \multirow{2}{*}{UniEval} & \multirow{2}{*}{BARTScore}\\
\cmidrule{2-7}
 & R-1 & R-2 & R-L & P & R & F1 \\
\midrule
LLaMa2-13B ZS & 27.65 & \underline{7.81} & 25.22 & \underline{85.30} & 84.73 &  85.01 &55.36 & \underline{76.03} & -4.78 \\
LLaMa2-13B OS & 26.53 & 7.01 & 24.39 & 84.86 & 84.43 & 84.65 & 54.94 & 71.98 & -4.81 \\
LLaMa2-13B DP & \underline{28.23} & 7.68 & \underline{25.83} & 85.18 & 85.12 & \underline{85.14} & 55.42 & 74.57 & \underline{-4.65} \\
LLaMa2-13B OSP & 25.84 & 6.66 & 23.67 & 84.51 & 84.55 & 84.53 & 54.65 & 69.23 & -4.74 \\
\midrule 
LLaMa2-70B ZS & 27.77 & 7.59 & 25.30 & 85.05 & 84.82 & 84.93 & 55.34 & \underline{74.06} & -4.73 \\
LLaMa2-70B OS & \underline{29.69} & \underline{8.49} & \underline{27.39} &  \underline{85.72} & \underline{85.49} & \underline{85.60} & \underline{55.63} & 71.46 & \underline{-4.48} \\
LLaMa2-70B DP & 28.74 & 8.06 & 26.29 & 85.31 & 84.98 & 85.14 & 55.49 & 72.36 & -4.67 \\
LLaMa2-70B OSP & 27.74 & 7.80 & 25.48 & 85.32 & 85.04 & 85.18 & 55.52 & 72.92 & -4.68 \\
\midrule
PaLM2 ZS & 27.95 & 8.95 & 25.99 & \underline{85.28} & 84.61 & 84.94 & 55.21 & 72.69 & -4.76\\
PaLM2 OS & \underline{28.81} & 9.05 & \underline{26.90} & 85.16 & \underline{84.71} & 84.93 & 55.35 & \underline{72.73} & \underline{-4.68}\\
PaLM2 DP & 28.77 & 9.13 & 26.65 & 85.27 & 84.66 & \underline{84.96} & \underline{55.31} & 72.41 & -4.75\\
PaLM2 OSP & 28.71 & \underline{9.34} & 26.67 & 85.14 & 84.61 & 84.87 & 55.28 & 72.72 & -4.74\\
\midrule
GPT-3.5 ZS & 26.60  & 6.30 & 24.36 & 85.57  & 84.68 & 85.12 & 55.47 & \textbf{81.31} & -4.75\\
GPT-4 ZS & 26.72 & 6.61 & 24.35 & 85.42 & 85.39 & 85.40 & 55.71 & 75.24 & -4.66\\
GPT-4 OS & 30.09 & 7.98 & 27.71 & 86.01 & 86.15 & 86.08 & 55.98 & 74.80 &-4.38 \\
GPT-4 OSP   & \textbf{31.47} & \textbf{8.62}  & \textbf{29.04} &  \textbf{86.19 } & \textbf{86.44}  & \textbf{86.31} & \textbf{56.04} & 75.55 &\textbf{-4.28} \\
\midrule
% GPT-4 OS+Wiki &  31.99 & 9.40 & 29.63 & 86.31 & 85.85 & 85.08 & 56.10 & 75.04 & -4.43 \\
*GPT-4 OS+IR & 31.96 & 9.43 & 29.60 & 86.27 & 85.88 & 86.07 & 56.44 & 78.56 & -4.38 \\
\bottomrule
\end{tabular}
\vspace{-2mm}
\caption{Automatic evaluation scores: we compare ROUGE, BERTScore, MoverScore, UniEval, and BARTScore on different settings. The superior scores among the same models are underlined, while the highest scores across all models and settings are highlighted in bold. * We use plugins including \hyperlink{https://www.whatplugin.ai/plugins/ab-web-search}{A\&B Web Search} and \hyperlink{https://www.keymate.ai/}{Keymate.ai}.}
\label{tab:auto_eval}
\end{table*}

\subsection{Evaluation Metrics}
\textbf{Automatic Evaluation} We evaluate the generated surveys using a range of automatic metrics including \textbf{ROUGE}, \textbf{BERTScore}~\cite{Zhang*2020BERTScore:}, \textbf{MoverScore}~\cite{zhao-etal-2019-moverscore}, \textbf{UniEval}~\cite{zhong-etal-2022-towards} and \textbf{BARTScore}~\cite{yuan2021bartscore}. Tab.~\ref{tab:auto_eval} provides an overview of results for the following LLMs: LLaMa2 (13B, 70B), PaLM2 (text-bison), GPT-3.5 (Turbo-0613) as well as GPT-4 (0613) across different prompt settings. We first notice that GPT-4 consistently outperforms other baselines, obtaining a significant improvement of around 2\% to 20\% when enhancing prompts. Specifically, \texttt{GPT-4 OSP} achieves the top spot under most situations. However, it is not to say that prompt enrichment always yields positive results. For instance, in the case of LLaMa2, one-shot and description prompts perform better than \texttt{OSP}. As for PaLM2, four types of prompts obtain similar results. 
% Following observations, we conduct further evaluations with the inclusion of information retrieval (\texttt{GPT-4 OS+IR)} and links to Wikipedia articles (Appendix~\ref{app:wiki}). 
When we add external knowledge (\texttt{GPT-4 OS+IR}), there is some improvement compared to \texttt{GPT-4 OS}. As our primary goal is to study the extent of knowledge LLMs possess in this task, we mainly focus on analyses in settings without external data. However, additional analysis about \texttt{GPT-4 OS+IR} setting can be found in Appendix~\ref{app:wiki}.
% However, our findings suggest that external knowledge does not manifestly enhance performance. Overall, it's remarkable to highlight that the LLMs can produce good-quality, domain-specific texts.

% , which is primarily attributed to LLaMa2's lower sensitivity to longer prompts.
% however, the \texttt{OSP} configuration achieves higher ranking in Rouge-2.
\textbf{Human and GPTs Evaluation} 
We employ two NLP experts, GPT-4 and G-Eval~\cite{liu-etal-2023-g} to evaluate surveys generated by the best \texttt{GPT-4 OSP} setting, focusing on 6 perspectives: \textbf{Readability}, \textbf{Relevancy}, \textbf{Hallucination}, \textbf{Completeness}, \textbf{Factuality}. Both GPT models and humans are required to score each aspect on a scale from $1$ to $5$, following the same guidelines. The detailed guidance can be found in Appendix~\ref{app:guide}. It's important to note that we implement a pre-selection stage in the choice of human experts (Appendix~\ref{app:guide}). Tab.~\ref{tab:human_gpt_score} shows that both human experts and GPTs agree that the generated surveys perform well across most aspects, though the \textit{completeness} exhibits marginally lowest scores. According to IAA, we can observe that human experts demonstrate a high consistent quality of the generated surveys whlie GPT-4 and G-Eval have more randomness. To better understand the degree of agreement between human experts and GPT-4 on ratings, we also calculate Kendall's $\tau$ and $p$-value as shown in Tab.~\ref{tab:Kendall's score}. We can observe that the \textit{Factuality} possesses the highest degree of correlation. In contrast, \textit{Redundancy} displays the lowest correlation while the other aspects exhibit relatively lower correlation levels. This difference is largely because \textit{Factuality} is based on objective ground truth, while \textit{Redundancy} is more dependent on subjective judgment. Notably, we can conclude that in most scenarios, GPT-4 showcases similar evaluative opinions as humans, despite showing a higher degree of variability across different independent sessions. Regarding \textbf{RQ1} and \textbf{RQ2}, we find that 1) LLMs can produce high-quality survey articles, and 2) with specific guidance, there's a strong consistency between GPT outputs and human judgment.

\begin{table*}[h]
    \centering
    \small
    \renewcommand{\arraystretch}{0.8}
    \begin{tabular}{cccccccc}
    \toprule
    &Evaluator & Readability & Relevancy & Redundancy & Hallucination & Completeness & Factuality \\
    \midrule
    % Mean\textsubscript{STD} \\
    % & Mean\textsubscript{STD} & Mean\textsubscript{STD} & Mean\textsubscript{STD} & Mean\textsubscript{STD} & Mean\textsubscript{STD} & Mean\textsubscript{STD} \\
    \multirow{3}{*}{Mean\textsubscript{STD}}
    & Human & $4.95_{0.30}$ & $4.88_{0.47}$ & $4.77_{0.53}$ & $4.84_{0.48}$ & $4.29_{0.68}$ & $4.80_{0.55}$
    \\
   & GPT-4 & $4.84_{0.32}$ & $4.67_{0.50}$ & $4.85_{0.34}$ & $4.86_{0.33}$ & $3.93_{0.42}$ & $4.56_{0.51}$ \\
   % & G-Eval & $4.90_{0.18}$ & $4.68_{0.55}$ & $4.31_{0.87}$& $4.94_{0.50}$& $4.22_{0.78}$ & $4.75_{0.81}$ \\
   & G-Eval & $4.77_{0.64}$ & $4.63_{0.68}$ & $4.27_{0.74}$& $4.94_{0.51}$& $4.26_{0.76}$ & $4.76_{0.66}$ \\
    \midrule
    % & IAA\textsubscript{\%} & IAA\textsubscript{\%} & IAA\textsubscript{\%} & IAA\textsubscript{\%} & IAA\textsubscript{\%} & IAA\textsubscript{\%} \\
    \multirow{3}{*}{IAA\textsubscript{\%}}
    & Human & $0.41_{96.96}$ & $0.47_{87.87}$ & $0.35_{68.68}$ & $0.41_{82.82}$ & $0.55_{66.66}$ & $0.59_{82.82}$ \\
   & GPT-4 & $0.09_{69.69}$ & $0.35_{64.64}$ & $0.003_{72.72}$ & $0.08_{75.75}$ & $0.32_{70.70}$ & $0.45_{63.63}$ \\
   & G-Eval & $0.06_{49.49}$ & $0.25_{38.38}$ & $0.01_{32.32}$ & $0.008_{95.95}$ & $0.42_{33.33}$ & $0.02_{58.58}$ \\
    \bottomrule
    \end{tabular}
    \vspace{-2mm}
    \caption{Human and GPTs Evaluation Results. We report the mean and standard deviation. We also quantify the IAA (inter-annotator agreement)~\cite{karpinska2021perils} between human experts and the GPT results, respectively, using Krippendorff's $\alpha$ coefficient and calculating the percentage (\%) of scores that are identical.}
    \label{tab:human_gpt_score}
    \vspace{-3mm}
\end{table*}

\begin{table}[h]
\small
\renewcommand{\arraystretch}{0.8}
    \centering
    \begin{tabular}{c|cc}
    \toprule
     % \multirow{2}{*}{}    &  \multicolumn{2}{c}{Kendall's} \\
     %     & $\tau$ & $p$ \\
    % \multirow{2}{*}{} & \multicolumn{2}{c}{GPT-4} & \multicolumn{2}{c}{G-Eval} \\
    % \cmidrule{2-5}
     &  $\tau$ & $p$ \\
         \midrule
         Readability & 0.16 & 0.09 \\
         Relevancy & 0.18 & 0.05 \\
         Redundancy & 0.07 & 0.46 \\
         Hallucination & 0.11 & 0.22 \\
         Completeness &  0.10 & 0.24 \\
         Factuality & 0.24 & 0.01 \\
    \bottomrule
    \end{tabular}
    \vspace{-2mm}
    \caption{The Kendall's $\tau$ correlation coefficient and $p$-value  between human and GPT-4.}
    \label{tab:Kendall's score}
    \vspace{-4mm}
\end{table}

\section{Analysis}
\vspace{-3mm}
In this section, we provide an in-depth analysis of the LLMs' internal knowledge on survey writing ability, and compare the evaluation scores of human and LLM assessments.

\begin{figure}
\centering
\begin{subfigure}{.35\textwidth}
  \centering
    \includegraphics[scale=0.11]{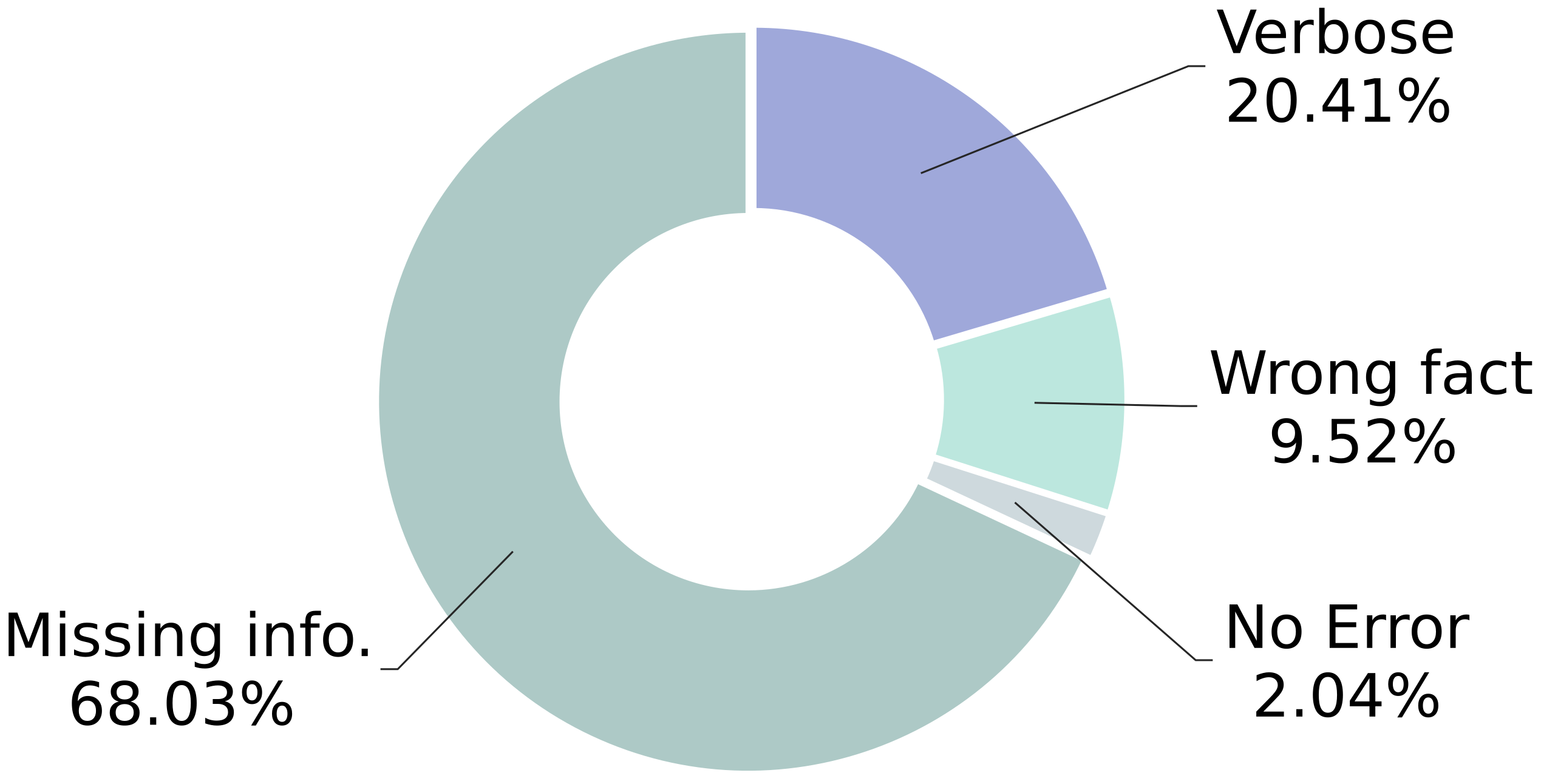}
  \caption{Error Type Distribution.}
  \label{fig:error_type}
\end{subfigure}

\begin{subfigure}{.35\textwidth}
  \centering
    \includegraphics[scale=0.11]{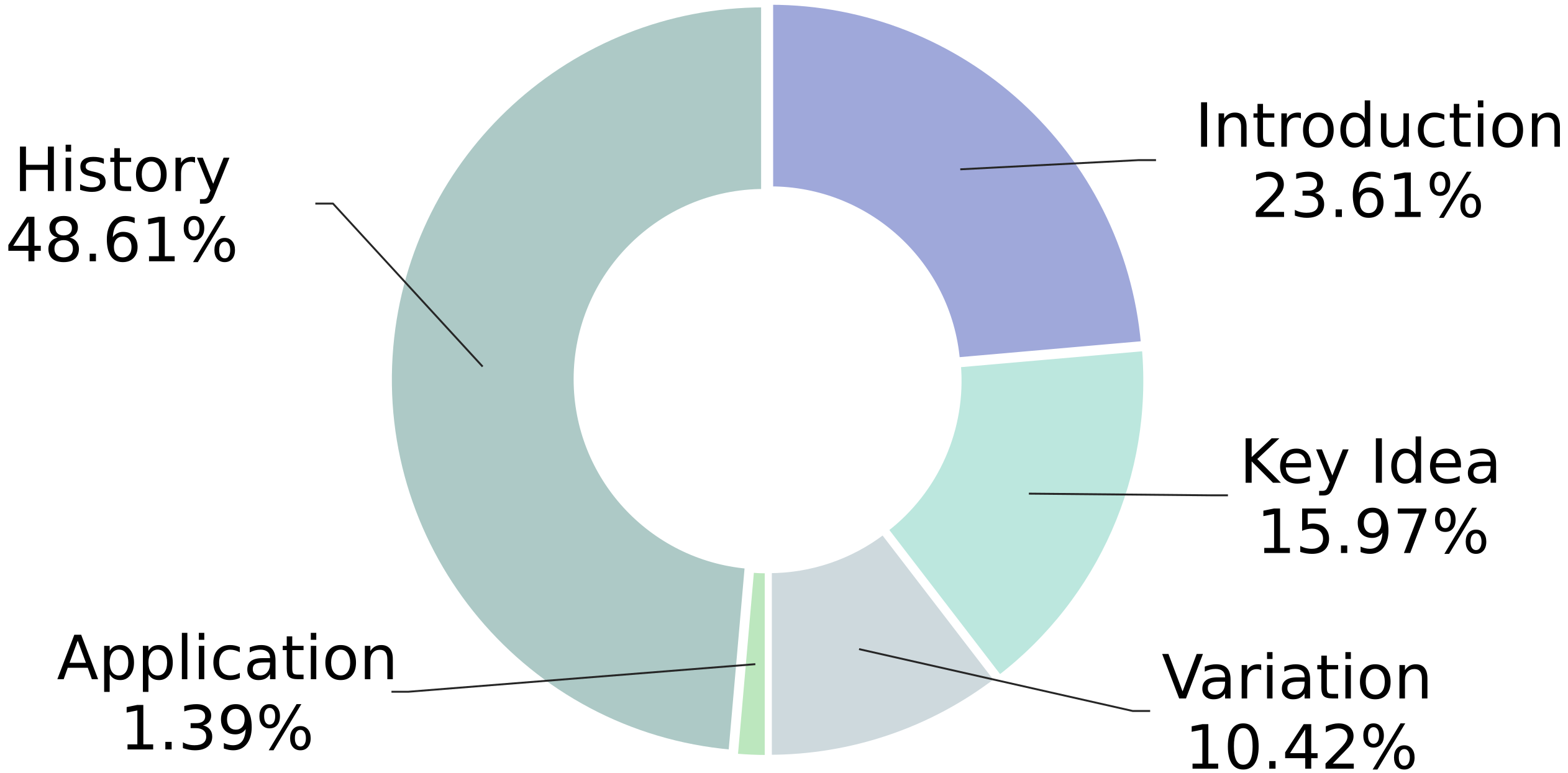}
  \caption{Section Error Distribution.}
  \label{fig:error_type}
\end{subfigure}
\vspace{-2mm}
\caption{Error Analysis by types and sections.}
\label{fig:combined_error_analysis}
\vspace{-2.5mm}
\end{figure}

% \subsection{Ability for Writing Surveys} 
\textbf{Error Types} We have shown that both automated and manual evaluations demonstrated that LLMs excel in crafting survey articles on scientific concepts. 
% Now we thoroughly analyze the best setting, \texttt{GPT-4 OSP}, and assess the errors present in the generated content, as identified by two human evaluators, and present a summary of error types and their distribution in Fig. ~\ref{fig:combined_error_analysis}. 
We analyze the best setting, \texttt{GPT-4 OSP}, assessing errors identified by two experts, and summarize error types and distributions in Fig. ~\ref{fig:combined_error_analysis}. We classify these errors into four categories: Verbose, Wrong Fact, Missing Information, and No Error (indicating flawless content). It shows that most errors are missing information, followed by verbosity and factual inaccuracies. Furthermore, the History and Introduction sections of the generated articles contained the highest number of errors, while the Application section exhibited the best. 
% We analyze it could be attributed to the complex nature of the History and Introduction sections, which typically require details such as origins, developments, and formation over the years by different scientists. These sections are required to be exactly accurate. Nevertheless, the Application is generally more broad, and reasonable explanations suffice.

\begin{figure}
  \centering
  \small{
\begin{tikzpicture}
    \begin{axis}[
        xbar, % Change ybar to xbar
        width=0.8\linewidth,
        height=3.8cm,
        xmax=0.75, % Change ymax to xmax
        xlabel={Percentage of Novel Entity Counts}, % Change ylabel to xlabel
        symbolic y coords={LLaMa2-13b, LLaMa2-70b, PaLM2, GPT-3.5, GPT-4}, % Change symbolic x coords to symbolic y coords
        ytick=data, % Change xtick to ytick
        nodes near coords,
        nodes near coords align={horizontal}, % Change vertical to horizontal
        y tick label style={rotate=0,anchor=east}, % Remove rotation
        bar width=0.2 cm,
        enlarge y limits=0.3, % Change enlarge x limits to enlarge y limits
        ylabel style={font=6pt}, % Change xlabel style to ylabel style
    ]
    \addplot[fill=color1, draw=none] coordinates {
        (0.6818309717,LLaMa2-13b) % Swap x and y coordinates
        (0.6242413611,LLaMa2-70b)
        (0.5224737529,PaLM2)
        (0.5995934554,GPT-3.5)
        (0.628388678,GPT-4)
    };
    \end{axis}

\end{tikzpicture}}
  \vspace{-3mm}
    \caption{Comparison of novel entity mentions.}
    \label{fig:entity}
      \vspace{-4mm}
\end{figure}
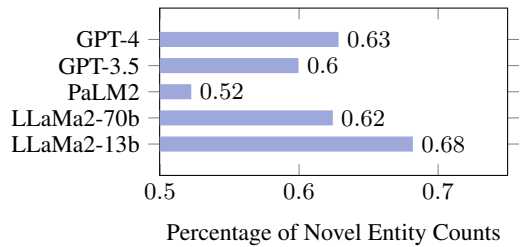

\textbf{Novel Entity Mention} To further investigate how interesting the generated content is, we look at the mentions of novel entities following \cite{10.1145/3491102.3502030}. Specifically, we examine the survey content by comparing the entities it contains with those in the ground truth.
% Specifically, we examine the survey content, juxtaposing the entities within it against those in the ground truth. 
We employ Stanza \cite{qi2020stanza} to identify all entities in both the LLM-generated text and the ground truth. Subsequently, we quantify the number of unique entities found in the LLM-generated content. For a fair comparison, we analyze the one-shot with prompt settings of LLaMa2-13b, PaLM2, and GPT-4, in addition to the ZS setting of GPT-3.5, as depicted in Fig.~\ref{fig:entity}. Our findings reveal that PaLM2 exhibited the least variation in entity mentions, while LLaMa2-13b showcased the most. Despite GPT-4's outstanding performance in both automated and human evaluations, we didn't discern a marked novelty in its entity mentions. We speculate that this might be an inherent compromise when generating high-fidelity content in relation to the ground truth. So far, regarding \textbf{RQ1}, although LLMs register commendable results based on predefined criteria, certain shortcomings are evident. Specifically, we observe some omitted details, particularly within the Introduction and History sections. While LLMs often introduce new entities, we don't find a significant correlation between this tendency and their performance. More case studies are in Appendix ~\ref{app:case}. 
% We present more case studies in Appendix ~\ref{app:case}. 

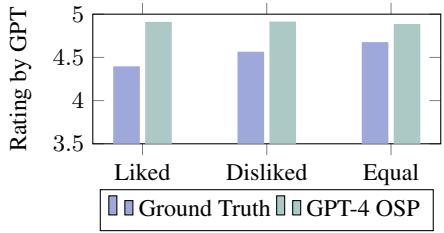
\begin{figure}[t]

  \centering
  \small
  \begin{tikzpicture}
    \begin{axis}[
      ybar,
      width=0.8\linewidth,
      height=3.3cm,
      ylabel={Rating by GPT},
      symbolic x coords={Liked, Disliked, Equal},
      xtick=data,
      % nodes near coords,
      % nodes near coords align={vertical},
      ymin=3.5,
      ymax=5,
      xlabel style={font=\small}, % Adjust font size for xlabel
      ylabel style={font=\small,yshift=-8pt}, % Adjust font size for ylabel
      xticklabel  style={font=\small}, 
      bar width=0.35 cm,
      enlarge x limits=0.2,
       legend style={
        at={(0.5,-0.35)}, % Adjust the position to below the x-axis
        anchor=north, % Align the legend to the north (top) side
        legend columns=-1, % Set to -1 to ensure the legend items are in one row
        font=\small,
      },
      %  legend style={
      %   at={(1.02,0.5)}, % Adjust the x-coordinate to move the legend
      %   anchor=west, % Align the legend to the west (right) side
      %   legend columns=1, % Set the number of legend columns to 2 for two rows
      %   font=\tiny,
      % },
    ]

    % Human-written bars in light blue
    \addplot[fill=color1,draw=none] coordinates {
      (Liked, 4.396551724)
      (Disliked, 4.565789474)
      (Equal, 4.676630435)
    };

    % GPT-written bars in light purple
    \addplot[fill=color2,draw=none] coordinates {
      (Liked, 4.910714286)
      (Disliked, 4.914473684)
      (Equal, 4.886363636)
    };

    \legend{Ground Truth, GPT-4 OSP}
    \end{axis}
  \end{tikzpicture}
  \vspace{-2mm}
  \caption{Evaluation comparison on ground truth and GPT-4 predictions, grouped by human preference.}
  \label{fig:h_vs_gpt}
  \vspace{-4mm}
  
\end{figure}

\textbf{LLM and Human Preference}
Previous studies have indicated that LLM-based evaluation methods tend to favor content generated by LLMs \cite{liu2023geval}. To test the veracity of this assertion within the context of survey generation tasks, we took the opportunity to investigate whether a similar observation holds in the context of survey generation tasks. Hence, we recruited two human experts in a blind side-by-side comparison of both the ground truth survey articles and articles generated using the best GPT-4 settings, and they assessed the content based on \lq Likeability\rq \cite{chiang-lee-2023-large}. Subsequently, we categorized the survey articles into three groups: a) (human experts) Liked, b) (human experts) Disliked, and c) Equal (equally good). The experts reached a significant agreement, reflected in a Cohen's Kappa score of 0.68 \cite{cohen1960coefficient}. In instances of disagreement, we randomly selected a score to reach a final consensus. We then apply the GPT-4 evaluation scores on the first four criteria except for \textit{Factuality} and \textit{Completeness} because both are impossible to do a blind test. We show the average ratings on all 99 concepts in Fig.~\ref{fig:h_vs_gpt}. One main observation is the bias of GPT-4 towards texts generated by itself and consistently conferring high ratings – an observation consistent with other studies \cite{liu2023geval}. When evaluating the ground truth, GPT-4 consistently assigns marginally lower ratings across all three categories. Intriguingly, GPT-4 shows a preference for the \textit{Disliked} group over the \textit{Liked} group when considering the ground truth, a tendency that diverges from human inclinations. This suggests that when assessing human-composed text, such as ground truth survey articles, GPT-4 might not yet be an impeccable substitute for human discernment. Thus, in response to \textbf{RQ3}, we found that GPT-4 exhibits a notable preference for machine-generated texts with specific biases. Furthermore, we contend that the complete replacement of human experts by GPT-4 is a challenging prospect. For instance, human expertise remains indispensable for manual content fact checking.
% it may be concluded that GPT-4 prefers machine-generated texts when given detailed evaluation instructions. However, there are still limitations. In the case of Factuality, human experts are needed to verify the content manually. 

\section{Discussion and Conclusion}
\vspace{-3mm}
In this work, we evaluate the ability of LLMs to write surveys on NLP concepts. We find that LLMs, particularly GPT-4, can author surveys following specific guidelines that rival the quality of human experts, even though there are shortcomings such as incomplete information. Our findings also indicate that GPT-4 may not be a perfect replacement for human judgment when evaluating human-composed texts, and certain biases exist when asking it to rate machine-generated texts. Nevertheless, the results imply that these advanced generative LLMs could play a transformative role in the realm of education. 
% We find that while GPT models can write convincing content, there are shortcomings, such as incomplete information. GPT-4 may not be a perfect replacement for human judgment when evaluating human-composed texts, and certain biases exist when asking it to rate machine-generated texts. 
% Our findings indicate that LLMs, particularly GPT-4, can author surveys following specific guidelines that rival the quality of human experts. This implies that these advanced generative LLMs could play a transformative role in the realm of education. 
They hold the promise of effectively structuring domain-specific knowledge tailored to general learners. This adaptability could potentially lead to a more interactive and personalized learning experience, enabling students to engage in query-driven studies that cater directly to their unique curiosities and learning objectives. 

\section*{Acknowledgements}
This work was partially supported by the Deutsche Forschungsgemeinschaft (DFG) through the priority program RATIO (SPP-1999), Grant No. 376059226. It was also supported by the Japan Society for the Promotion of Science (JSPS) KAKENHI, Grant Number 24K20832. We would like to thank these funding agencies for their generous support. Additionally, we would like to thank the anonymous reviewers for their helpful feedback.

\section*{Limitations and Ethical Considerations}
GPT-4 can generate contemporary, accessible content, but sometimes compromises depth and detail, leading to potential information gaps and occasional factual inaccuracies. This requires extra verification. Comparing ratings between human experts and GPT-4 revealed a systematic bias in GPT evaluations, which can skew outcomes and mislead quality perception.

The primary objective of this work is to explore the potential applicability of LLMs in the field of education, with a specific emphasis on enhancing the understanding of LLMs' generative capabilities within computer science. Our focus is on assessing the efficacy and identifying the boundaries of the generated texts. It's important to note that the generated texts are devoid of any harmful content, and all data used and produced in this study contains no private information.

\bibliography{anthology,custom}
\bibliographystyle{acl_natbib}

\clearpage
\newpage
\appendix
\onecolumn

\section{Human Evaluation Guidance}
\label{app:guide}

The detailed human evaluation guidance is listed in the following: 
\begin{enumerate}
  \item \textbf{Readability}:
    \begin{itemize}
      \item \textbf{1 (bad):} The text is highly difficult to read, full of grammatical errors, and lacks coherence and clarity.
      \item \textbf{5 (good):} The text is easy to read, well-structured, and flows naturally.
    \end{itemize}

  \item \textbf{Relevancy}:
    \begin{itemize}
      \item \textbf{1 (bad):} The generated text is completely irrelevant to the given context or prompt.
      \item \textbf{5 (good):} The generated text is highly relevant and directly addresses the given context or prompt.
    \end{itemize}

  \item \textbf{Redundancy}:
    \begin{itemize}
      \item \textbf{1 (bad):} The text is excessively repetitive, containing unnecessary repetitions of the same information. For example, each section should have 50-150 tokens. If it is too long, we should give a low rating.
      \item \textbf{5 (good):} The text is concise and free from redundancy, providing only essential information.
    \end{itemize}

  \item \textbf{Hallucination}:
    \begin{itemize}
      \item \textbf{1 (bad):} The generated text includes false or misleading information that does not align with the context or is factually incorrect.
      \item \textbf{5 (good):} The generated text is free from hallucinations and provides accurate and contextually appropriate information.
    \end{itemize}

  \item \textbf{Completeness/Accuracy}:
    \begin{itemize}
      \item \textbf{1 (bad):} The generated text is incomplete (missing key information), leaving out crucial details or providing inaccurate information.
      \item \textbf{5 (good):} The generated text is comprehensive, accurate, and includes all relevant information.
    \end{itemize}

  \item \textbf{Factuality}:
    \begin{itemize}
      \item \textbf{1 (bad):} The text contains a significant number of factual inaccuracies or false statements, especially in History and Main Idea. For example, Year or people are wrong.
      \item \textbf{5 (good):} The text is factually accurate, supported by evidence, and free from misinformation.
    \end{itemize}

\end{enumerate}

\textbf{Pre-selection} We initially engaged four NLP specialists to assess the surveys produced by GPT on 20 handpicked topics, as listed in Tab.~\ref{tab:concepts}. The evaluation scores across four model configurations are showcased in Tab.~\ref{tab:human-20-eval}. Noting the considerable standard deviations among the evaluations of the four judges, we subsequently opted for two judges with a higher alignment in their scores to assess the entirety of the concepts.

\begin{table}[t]
\centering
\begin{tabular}{lll}
\toprule
BERT & Autoencoders & Clustering \\
Decision Trees & Ensemble Learning & Gaussian Mixture Model \\
Generative Adversarial Network & Gradient Boosting & Hidden Markov Models \\
Knowledge Graphs & Language Modeling & Long Short-Term Memory Network \\
Maximum Marginal Relevance & Meta Learning & Multilingual BERT \\
Perceptron & Relation Extraction & Residual Neural Network \\
RMSprop Optimizer & Sentiment Analysis & \\
\bottomrule
\end{tabular}
\vspace{1mm}
\caption{The 20 selected concepts in pre-selection stage. }
\label{tab:concepts}
\end{table}

\begin{table}[h]
\small
\centering
\begin{tabular}{ccccccc}
\toprule
     \multirow{2}{*}{Model} & Readability & Relevancy & Redundancy & Hallucination & Completeness & Factuality  \\
     & Mean\textsubscript{STD} & Mean\textsubscript{STD} & Mean\textsubscript{STD} & Mean\textsubscript{STD} & Mean\textsubscript{STD} & Mean\textsubscript{STD} \\
     \midrule
     GPT-3.5 ZS & $4.01_{0.98}$ & $3.66_{1.61}$ & $3.62_{1.04}$ & $3.82_{1.18}$ & $2.77_{0.94}$ & $3.56_{0.83}$ \\
     GPT-4 ZS & $4.56_{0.65}$ & $4.25_{0.76}$ & $4.20_{0.69}$ & $4.52_{0.79}$ & $3.50_{0.71}$ & $3.91_{0.92}$ \\
     GPT-4 ZPS\tablefootnote{ZPS means zero-shot with description prompt.} & $4.58_{0.72}$ & $4.41_{0.75}$ & $4.03_{0.81}$ & $4.56_{0.64}$ & $3.93_{0.69}$ & $4.07_{0.93}$ \\
     GPT-4 OPS & $4.60_{0.60}$ & $4.35_{0.79}$ & $4.20_{0.64}$ & $4.45_{0.78}$ & $3.90_{0.70}$ & $4.96_{1.07}$ \\
     \bottomrule
\end{tabular}
\caption{Human evaluation scores on 20 topics of four human experts.}
\label{tab:human-20-eval}
\end{table}

% \textcolor{red}{add the 20 concept results: pre-selection, etc...}

\section{Comparisons with External Knowledge}
\label{app:wiki}
We conduct further evaluations with the inclusion of links to Wikipedia articles (\textit{GPT-4 OS+Wiki)} and information retrieval (\texttt{GPT OS+IR}). In the \texttt{GPT-4 OS+Wiki} set up, we apply Embedchain~\cite{embedchain} which supports embedding open sources for LLM querying. We crawl Wikipedia articles to concepts in Surfer 100 datasets, yielding 87 effective links. We then prompt GPT-4 to generate survey articles for the respective 87 topics, providing corresponding Wikipedia links and a sample survey as references. As for the setting \texttt{GPT-4 OS+IR}, we ask GPT-4 to \textit{'Search on the web for helpful information'} in the prompt and utilitze the web search  APIs. Table~\ref{tab:wiki_eval} shows the comparison results between the GPT-4 with and without external knowledge. It's clear to see that both Wikipedia links and the information retrieval component significantly improve the Rouge scores. Notably, searching for web sources efficiently improve both the MoverScore and UniEval. In summary, external knowledge aids GPT-4 in generating higher-quality survey articles than using internal knowledge only. This suggests that LLMs possess limited proficiency when functioning as an academic search engine.

\begin{table*}[t]
\centering
\small
\renewcommand{\arraystretch}{0.8}
\begin{tabular}{lccccccccc}
\toprule
\multirow{2}{*}{} & \multicolumn{3}{c}{ROUGE} & \multicolumn{3}{c}{BERTScore} & \multirow{2}{*}{MoverScore} & \multirow{2}{*}{UniEval} & \multirow{2}{*}{BARTScore}\\
\cmidrule{2-7}
 & R-1 & R-2 & R-L & P & R & F1 \\
\midrule
GPT-4 OS & 30.09 & 7.98 & 27.71 & 86.01 & \textbf{86.15} & \textbf{86.08} & 55.98 & 74.80 & \textbf{-4.38} \\
GPT-4 OS+Wiki &  \textbf{31.99} & 9.40 & \textbf{29.63} & \textbf{86.31} & 85.85 & \textbf{86.08} & 56.10 & 75.04 & -4.43 \\
GPT-4 OS+IR & 31.96 & \textbf{9.43} & 29.60 & 86.27 & 85.88 & 86.07 & \textbf{56.44} & \textbf{78.56} & \textbf{-4.38} \\
\bottomrule
\end{tabular}
% \vspace{-2mm}
\caption{Automatic evaluation scores for settings with external knowledge. Highest scores are highlted.}
\label{tab:wiki_eval}
\end{table*}

\section{More Case Study and Observations}
\label{app:case}
\subsection{Understanding of ``Survey''}

\begin{figure}[t]
  \centering
  \adjustbox{width=\textwidth}{\includegraphics{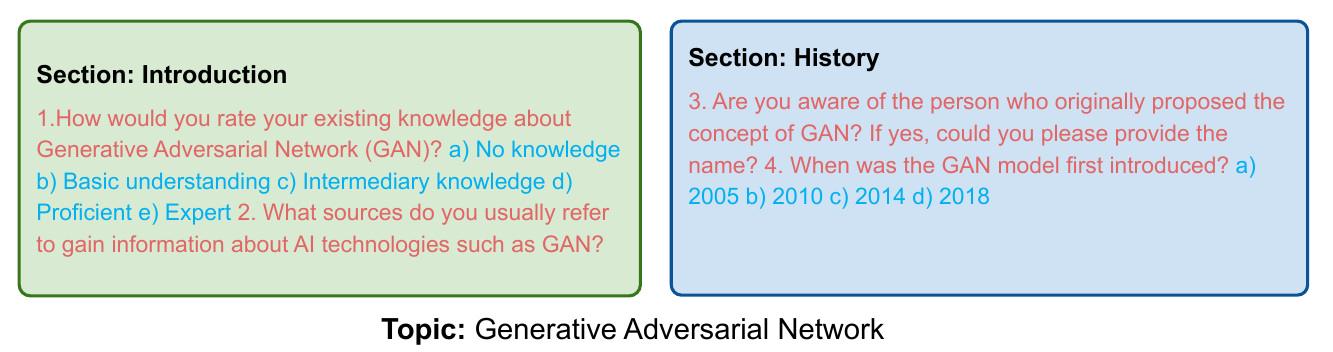}}
  \caption{An example of generated questionnaire as a survey (part).}
  \label{tab: survey_term}
\end{figure}

\begin{figure}[h!]
  \centering
  \adjustbox{width=0.95\textwidth}{\includegraphics{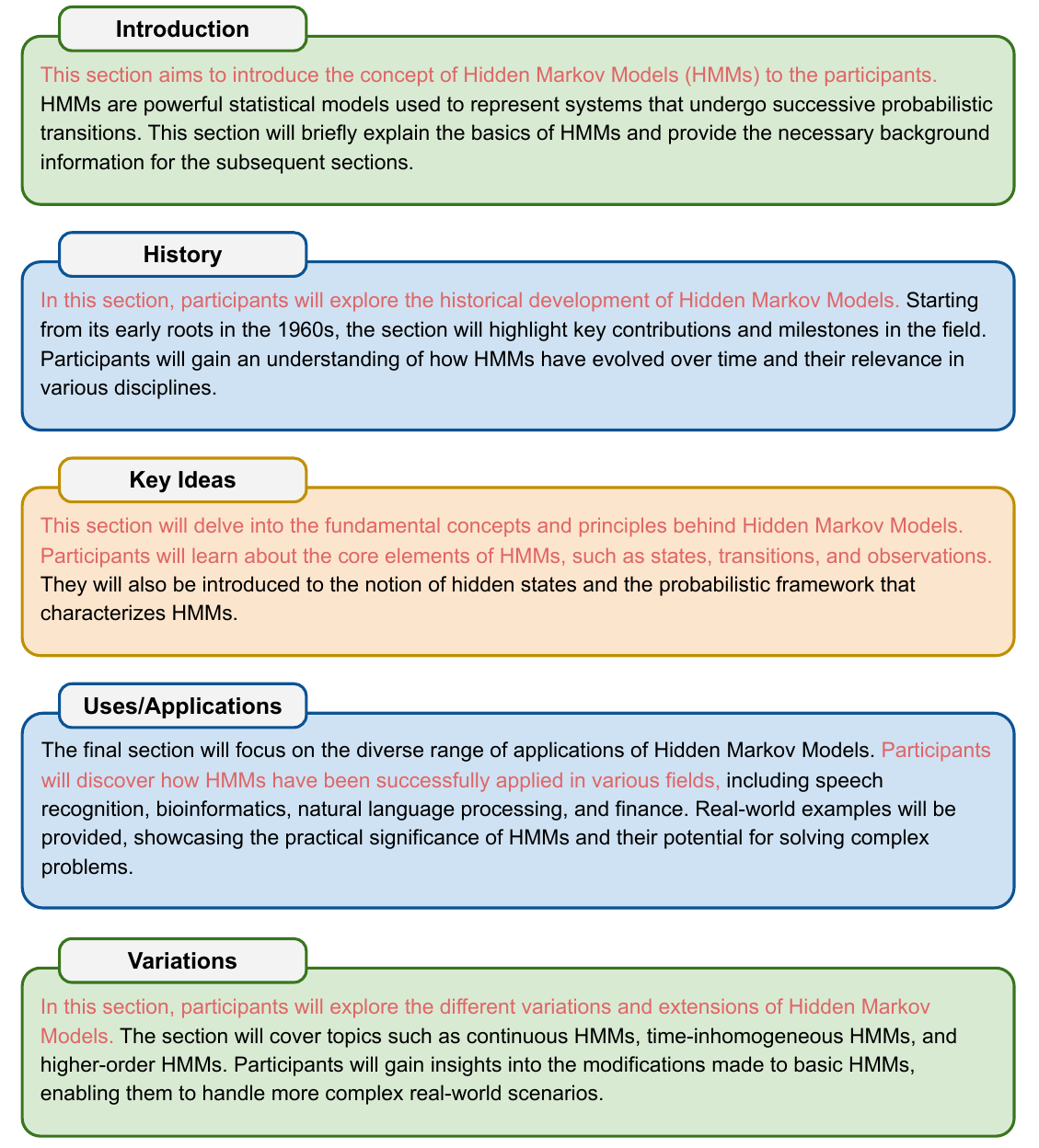}}
  \caption{An example of an invalid generated survey.}
  \label{tab: invalid}
\end{figure}

When we give the prompt to GPT models by asking them to write a ``survey'', they sometimes generate survey articles as desired, but they will write other types of content. For example, as indicated in Fig.~\ref{tab: survey_term}, it appears that GPT would understand the term ``Survey'' as the questionnaire. Moreover, even if they are able to generate a survey article in the format, there is still the situation that the generated content is not a typical survey. As shown in Fig. \ref{tab: invalid}, there are inconsequential sentences in an attempt to extend and explain the provided text. For example, it repeats saying \textit{this section} and \textit{participants}. But this is mostly observed in the GPT-3.5 zero-shot setting. 

% \begin{table}[t]
% \begin{tabularx}{\textwidth}{X}
% \hline
% \toprule
% \textbf{Topic:} Generative Adversarial Network \\ \midrule
% \textbf{Section:} Introduction

% 1. How would you rate your existing knowledge about Generative Adversarial Network (GAN)?
%     a) No knowledge
%     b) Basic understanding
%     c) Intermediary knowledge
%     d) Proficient
%     e) Expert
% 2. What sources do you usually refer to gain information about AI technologies such as GAN?\\
% \midrule
% \textbf{Section:} History

% 3. Are you aware of the person who originally proposed the concept of GAN? If yes, could you please provide the name?
% 4. When was the GAN model first introduced?
%     a) 2005
%     b) 2010
%     c) 2014
%     d) 2018 \\
% \bottomrule
% \end{tabularx}
% \vspace{1mm}
% \caption{An example of generated questionnaire as qurvey (part). }
% \label{tab: survey_term}
% \end{table}

\subsection{Incomplete Information}
\begin{figure}
  \centering
  \adjustbox{width=\textwidth}{\includegraphics{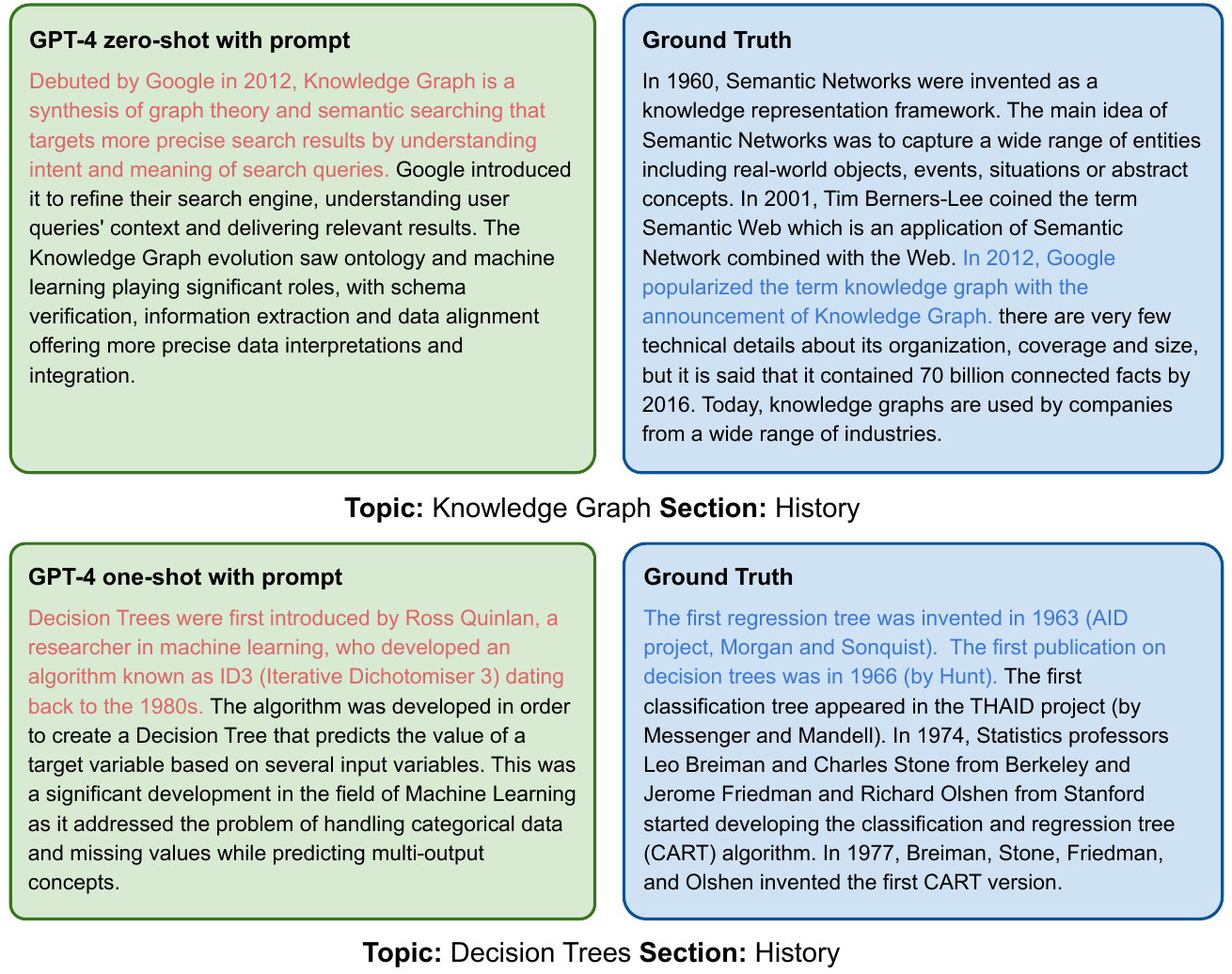}}
  \caption{Two example surveys with incomplete information.}
  \label{tab: incomplete}
\end{figure}

In the ``History'' section, GPT models occasionally produce incomplete evolutionary history, and thus, potentially result in misleading information. For instance, in Fig.~\ref{tab: incomplete}, when discussing the Knowledge Graph topic, GPT-4 model simply asserts that the term was invented by Google, while the reality is that the concept of Knowledge Graph has a long history, and it is Google that popularized the term. Similarly, in the case of the topic on Decision Trees, although the GPT model yields accurate context, it ignores landmark events and consequently causes misunderstandings.

\subsection{Nebulous Sentence Structure}
\begin{figure}
  \centering
  \adjustbox{width=\textwidth}{\includegraphics{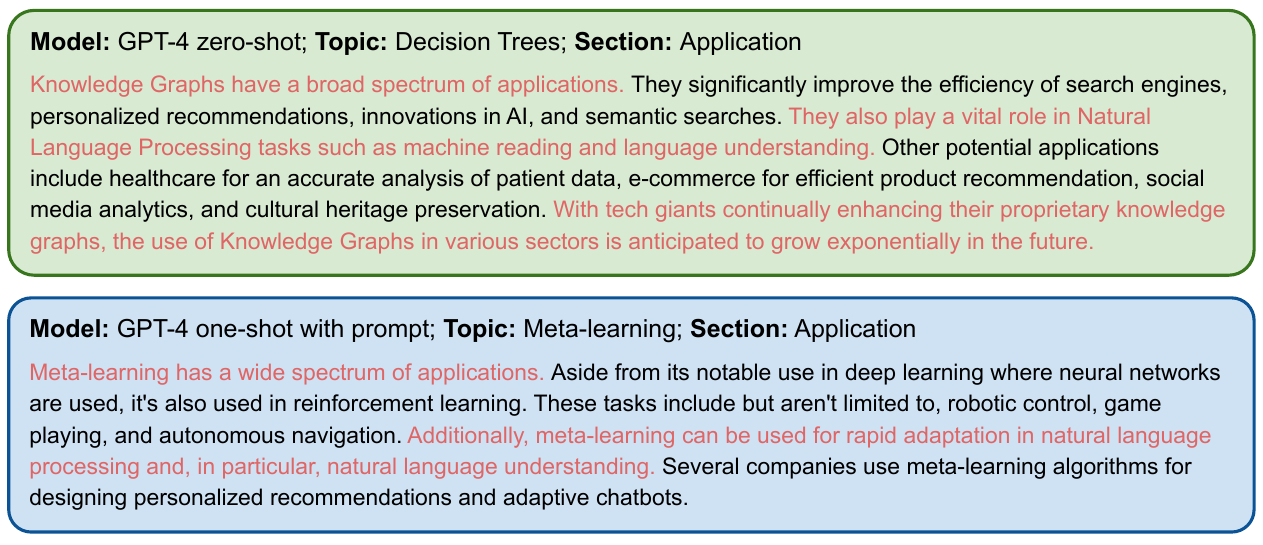}}
  \caption{Example surveys with nebulous sentence structure.}
  \label{tab: vague}
\end{figure}

We observe that GPT models frequently construct sentences, especially within the ``Application'' Section, that employ a rather vague sentence structure, which lacks specificity and can be used in different NLP topics. As shown in Fig. \ref{tab: vague}, it is evident that GPT models tend to generate similar sentences, such as ``The Topic has a wide spectrum of applications'' and ``The Topic plays a vital role in Natural Language Processing and Natural Language Understanding''; These statements hold significant meaning when ``The Topic'' is substituted with any NLP topics.

\subsection{High-quality Survey}
\begin{figure}
  \centering
  \adjustbox{width=\textwidth}{\includegraphics{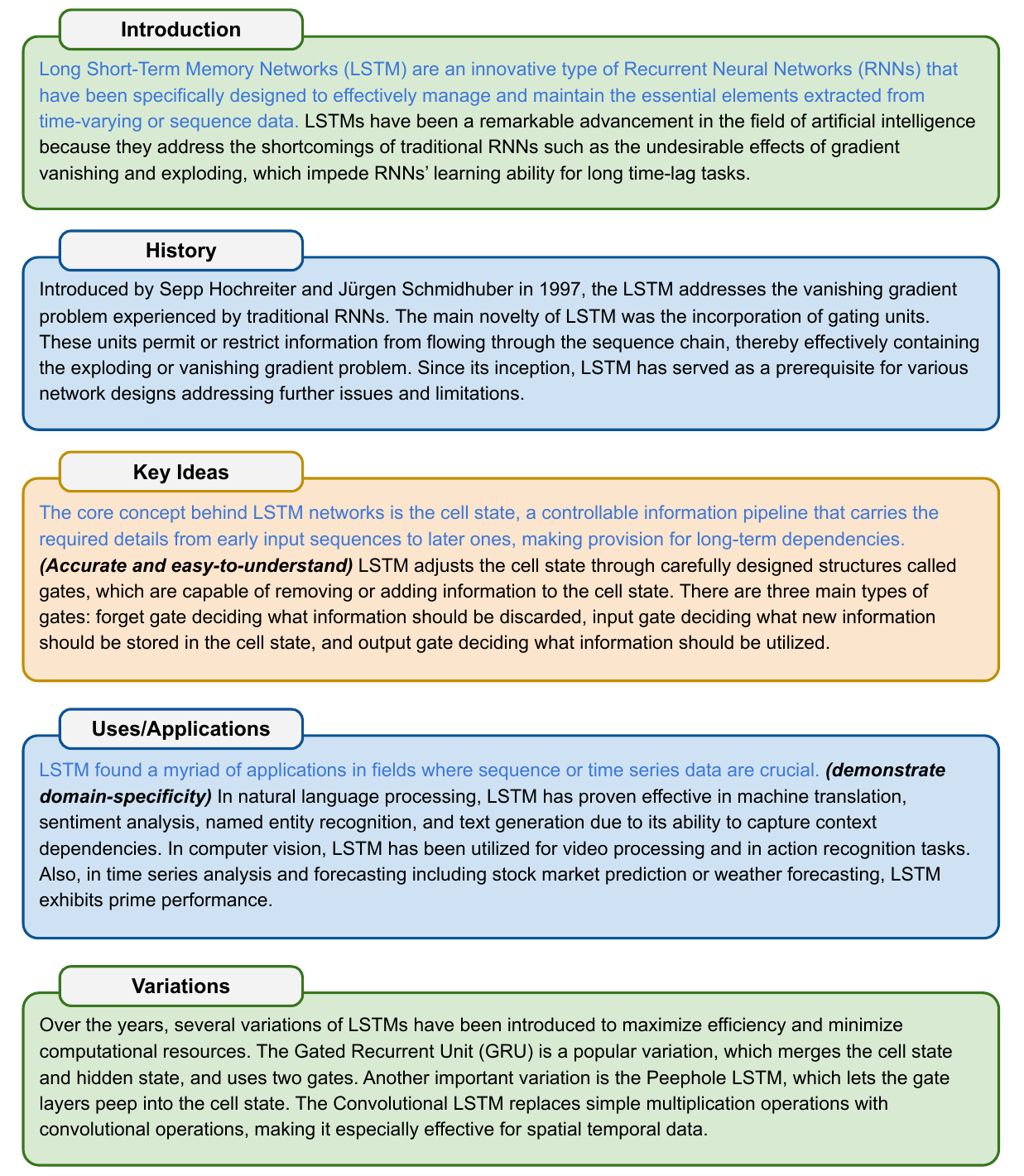}}
  \caption{An example of the high-quality survey.}
  \label{tab: high-quality}
\end{figure}

We also present a high-quality generated survey in Fig~\ref{tab: high-quality}. It is designed to read and understand easily, providing readers with comprehensive and detailed information. The example survey on LSTM is well-structured, with a summary provided in the first sentence and followed by the detailed explanation in each section. Specially, when discussing applications, it demonstrates a high level of domain specificity. Most importantly, the generated information is both accurate and concise.

\subsection{Going Beyond the Ground Truth}
\begin{figure}[t]
  \centering
  \adjustbox{width=\textwidth}{\includegraphics{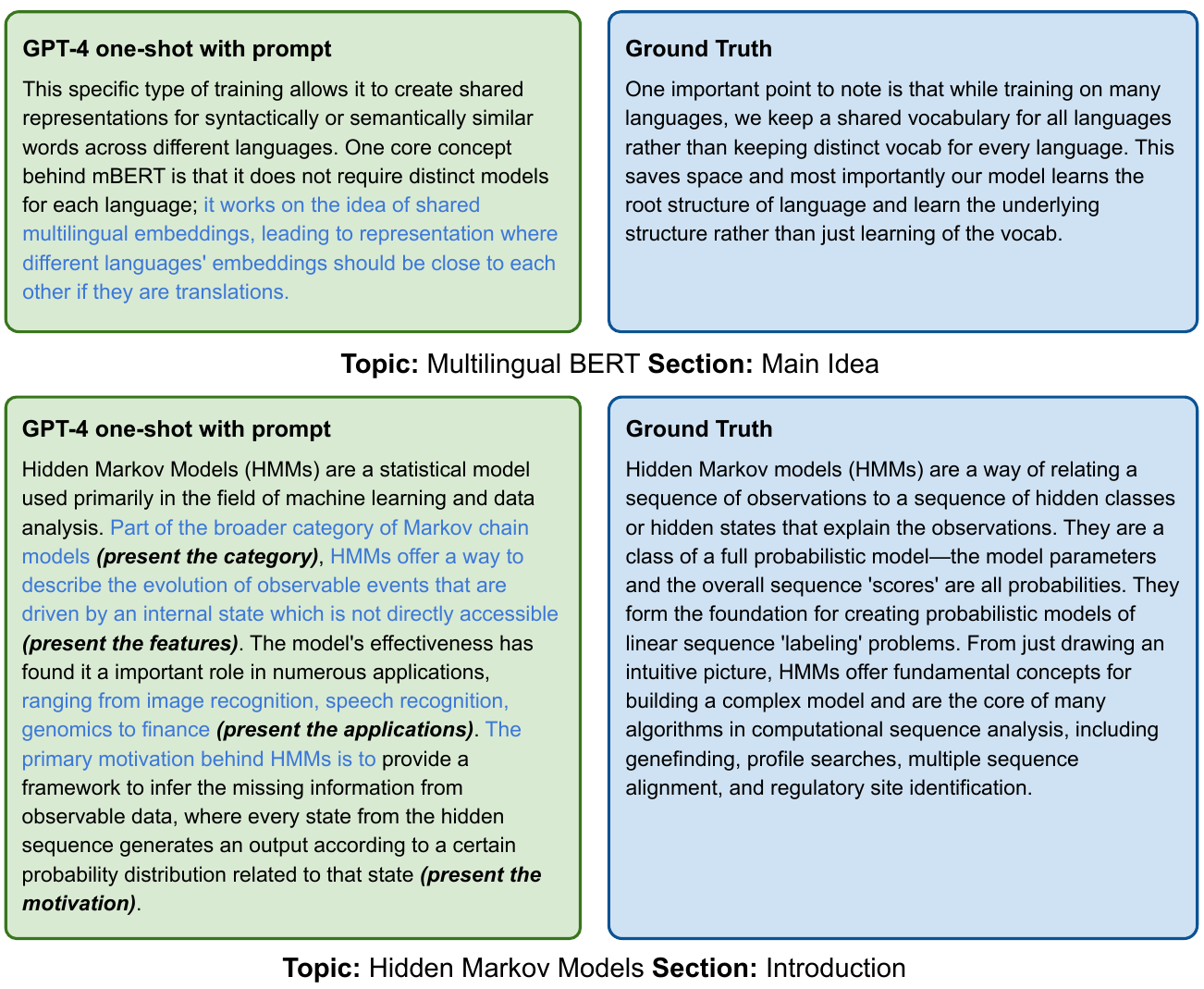}}
  \caption{Two examples showing that the generated output is better than the ground truth.}
  \label{tab: high-quality1}
\end{figure}

GPT-4 based methods maintain an overall high-quality response regarding all aspects. We show two examples by comparing the \textit{GPT-4 one-shot with prompt} setting result with the ground truth in Fig.~\ref{tab: high-quality1}. In the first topic, multilingual BERT (mBERT), GPT successfully points out that the key idea behind mBERT is mapping words from distinct languages into a shared embedding space. However, the ground truth only mentions shared vocabulary, which is superficial. In the second example, Hidden Markov Models (HMMs), the GPT response is more precise and more complete than the ground truth. One can find that the content flow is present as algorithm category $\rightarrow$ features $\rightarrow$ applications $\rightarrow$ motivation (highlighted in bold and italicized words). In contrast, the ground truth texts spend a lot of words to which category HMM belongs, including many terminologies which is less informative. 

\textbf{Limitations of Ground Truth} We refer to this previous work \cite{li-etal-2022-surfer100} on how the ground truth was generated. In general, the human writer was asked to rely on web data when writing the survey article; while these data were collected in the year 2021, it may be hard to say if it is a fair ROUGE score comparison with GPT models in Tab.~\ref{tab:auto_eval}. While the ground truth may not be a perfect reference, in this work, we focus more on human evaluation and case studies. 

% One may find that when applying the plugin, it reaches the best results, and the reason might be that it searches the web data, which may be close to how ground truth was created. 

\section{Potential Risks}
Sole dependence on LLMs for educational content can lead to a homogenization of information and lack the nuanced understanding that human experts bring. If not properly vetted, the occasional factual errors made by GPT-4 could lead to the propagation of misinformation, especially detrimental in an educational context. Over time, heavy reliance on automated systems might diminish the role of human experts in content creation, leading to a potential loss of rich, experience-based insights. The observed systematic bias in GPT evaluations can lead researchers to draw incorrect conclusions about the quality of content, potentially impacting future research and educational endeavors.

\section{Experimental Details}
In our work, we mainly utilize the paid GPT-4 model to generate Wiki-style survey articles and further explore its capabilities to score the generations. The total cost of these experiments is around 230 USD. During the automatic evaluation stage, we compute the ROUGE score and BERTScore using the officially provided APIs: \texttt{rouge}\footnote{https://pypi.org/project/rouge/} and \texttt{bert\_score}\footnote{https://github.com/Tiiiger/bert\_score}. For calculations involving MoverScore, UniEval, and BARTScore, we directly download their source codes. All experiments were performed using the high-performance machine with 4 A100 40GB NVIDIA cards. As the experiments do not involve fine-tuning, for each setting, we were able to finish in a few hours.  
As for the human evaluation stage, we calculate the Krippendorff's and Kendall's scores with the authorized APIs \texttt{krippendorff}\footnote{https://pypi.org/project/krippendorff/} and \texttt{scipy}\footnote{https://scipy.org/}.

\end{document}